\documentclass[sigconf]{acmart}
\AtBeginDocument{%
  }

\setcopyright{acmlicensed}
\copyrightyear{2026}
\acmYear{2026}
\setcopyright{cc}
\setcctype{by}
\acmConference[ICAIF '26]{7th ACM International Conference on AI in Finance}{November 14--17, 2026}{Milan, AA, Italy}
\acmBooktitle{7th ACM International Conference on AI in Finance (ICAIF '26), November 14--17, 2026, Milan, AA, Italy}
\acmDOI{10.1145/3858063.3858683}
\acmISBN{979-8-4007-3087-0/2026/11}

\usepackage[utf8]{inputenc}
\usepackage{url}
\usepackage{amsfonts}
\usepackage{multirow}

\begin{document}

\title{Mind the Shift: Decoding Monetary Policy Stance from FOMC Statements with Large Language Models}

\author{Yixuan Tang}
\orcid{0009-0006-2405-2026}
\affiliation{%
  \institution{The Hong Kong University of Science and Technology}
  \city{Hong Kong}
  \country{China}}
\email{ytangch@connect.ust.hk}

\author{Yi Yang}
\orcid{0000-0001-8863-112X}
\affiliation{%
  \institution{The Hong Kong University of Science and Technology}
  \city{Hong Kong}
  \country{China}}
\email{imyiyang@ust.hk}

\renewcommand{\shortauthors}{Tang and Yang}

\begin{abstract}
Federal Open Market Committee (FOMC) statements are a major source of monetary-policy information, and even subtle changes in their wording can move global financial markets. A central task is therefore to measure the hawkish--dovish stance conveyed in these texts. Existing approaches typically treat stance detection as a standard classification problem, labeling each statement in isolation. However, the interpretation of monetary-policy communication is inherently relative: market reactions depend not only on the tone of a statement, but also on how that tone shifts across meetings. We introduce \textbf{Delta-Consistent Scoring (DCS)}, an annotation-free framework that maps frozen large language model (LLM) representations to continuous stance scores by jointly modeling absolute stance and relative inter-meeting shifts. Rather than relying on manual hawkish--dovish labels, DCS uses consecutive meetings as a source of self-supervision. It learns an absolute stance score for each statement and a relative shift score between consecutive statements. A delta-consistency objective encourages changes in absolute scores to align with the relative shifts. This allows DCS to recover a temporally coherent stance trajectory without manual labels. Across four LLM backbones, DCS consistently outperforms supervised probes and LLM-as-judge baselines, achieving up to 71.1\% accuracy on sentence-level hawkish--dovish classification. The resulting meeting-level scores are also economically meaningful: they correlate strongly with inflation indicators and are significantly associated with Treasury yield movements. Overall, the results suggest that LLM representations encode monetary-policy signals that can be recovered through relative temporal structure.

\end{abstract}





\begin{teaserfigure}
  \centering
  \includegraphics[width=.88\linewidth]{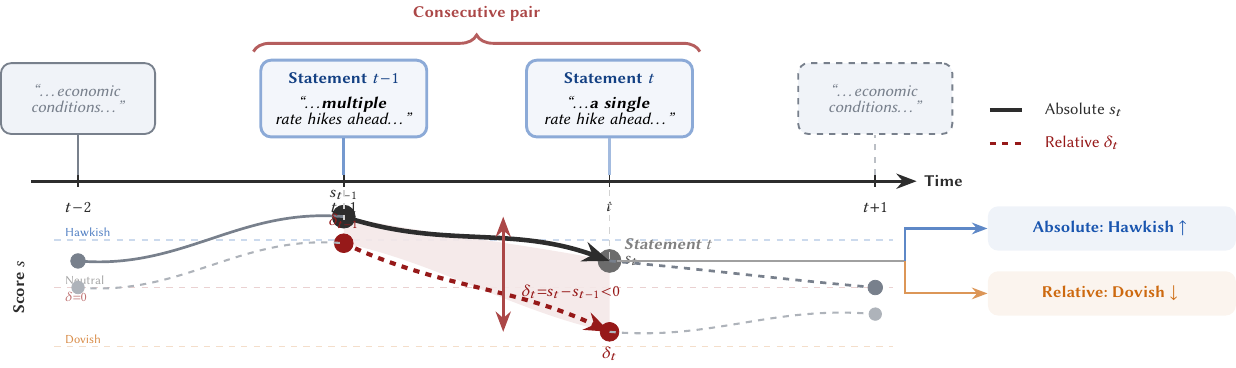}
  \caption{Monetary-policy stance is inherently relative. Statement $t$ (``a single rate hike ahead'') is hawkish in isolation ($s_t$ above neutral), yet signals a dovish shift relative to Statement $t{-}1$ (``multiple rate hikes ahead'').}
  \Description{A timeline shows four consecutive meetings. The highlighted pair changes from multiple rate hikes at the previous meeting to a single rate hike at the current meeting. Below the timeline, the absolute stance score decreases but remains above neutral, while the relative shift is negative. Two callouts identify the current statement as hawkish in absolute stance and dovish relative to the previous statement.}
  \label{fig:motivation}
\end{teaserfigure}

% \received{20 February 2026}

\maketitle

\section{Introduction}

The Federal Open Market Committee (FOMC) communicates U.S.\ monetary policy decisions through statements released after its policy meetings, and these texts move global financial markets~\citep{HansenStephen2018TADW, GorodnichenkoYuriy2023TVoM, shah-etal-2023-trillion}
These texts encode the Fed's monetary-policy stance: a hawkish statement signals a preference for higher interest rates to contain inflation, while a dovish statement signals a preference for lower rates to support economic growth. Because market participants parse these statements to form expectations about the future path of interest rates~\citep{LuccaDavidO2009MCBC}, even subtle changes in language can trigger large market reactions. For example, when Fed Chair Jerome Powell delivered an 8-minute speech in August 2022 signaling a tightening stance, U.S.\ equity markets lost nearly \$3 trillion in value that day, followed by over \$6 trillion in losses over the next three days~\citep{shah-etal-2023-trillion}.

Given its direct economic consequences, measuring the hawkish-dovish stance of Fed statements is critical. Yet the task remains challenging. Traditional dictionary-based methods~\citep{LuccaDavidO2009MCBC, LOUGHRANTIM2011WIaL} rely on keyword counting and predetermined word lists, ignoring the discourse-level context that gives policy language its meaning. Supervised approaches~\citep{shah-etal-2023-trillion, Christiano2025From} address this by training classifiers on expert-annotated sentences, but such annotations are labor-intensive, inherently subjective, and prone to degradation as policy language evolves across different rate cycles~\citep{alma991001481599703412}. While recent LLM-as-judge methods~\citep{hansencangpt2023, geiger2025monetary} reduce the reliance on manual annotation, they remain highly sensitive to prompt design and decoding parameters, and their outputs can be difficult to reproduce. While these approaches differ in supervision and modeling assumptions, they share a deeper limitation.

They treat stance detection as an isolated, absolute classification task, labeling individual statements without capturing the sequential structure of FOMC statements. Yet financial markets react not merely to the absolute stance of a statement, but to how that stance departs from prior statements~\citep{LuccaDavidO2009MCBC, DohTaeyoung2020DFRC}. Treasury yield levels are driven not only by the rate decision itself, but also by relative shifts in policy communication~\citep{GurkaynakRefetS2005DASL}. Figure~\ref{fig:motivation} illustrates a simple example: a moderately hawkish statement can still imply a dovish shift when it follows a more strongly hawkish statement. Stance is therefore relative, and any approach that ignores the inter-meeting trajectory discards a first-order signal.

This observation suggests that stance should be recovered from the temporal relations between consecutive statements rather than from isolated texts alone. To do so without manual labels, we turn to representation probing, which has shown that pretrained LLM representations contain semantic information that can be extracted with lightweight modules~\citep{ccs2023, park2025steer, zou2025representation}. We hypothesize that these latent representations also encode information about the hawkish-dovish policy spectrum, and that the relative shifts between consecutive meetings provide a natural supervision signal for extracting it.

In this paper, we propose \textbf{Delta-Consistent Scoring (DCS)}, an annotation-free framework that learns to map frozen LLM representations to continuous hawkish-dovish scores. Rather than relying on human annotations, it exploits the consecutive nature of FOMC meetings to construct a learning signal. 
% Specifically, we first encode a sequence of FOMC statements into LLM representations, and then train a lightweight dual-axis projection over these representations. 
% One axis produces an absolute stance score for each statement, while the other estimates the relative shift between consecutive meetings. We jointly optimize both axes by enforcing a delta-consistency constraint: the difference between the absolute scores of two consecutive statements should align with the relative shift estimated for the same pair. 
Specifically, we first encode a sequence of FOMC statements into LLM representations, and then train a lightweight scoring module over these representations. It learns an absolute stance score for each statement and a relative shift score between consecutive meetings. We tie these two signals together with a delta-consistency constraint, which encourages the change in absolute stance between two consecutive statements to match the estimated relative shift for the same pair.
This constraint turns the temporal ordering of FOMC statements into a structured source of self-supervision.

We evaluate DCS across different LLMs ranging from 1B to 14B parameters. Although DCS requires no stance labels during training, we benchmark it against labeled test data following the evaluation protocol of \citet{shah-etal-2023-trillion}. Our method consistently outperforms both supervised baselines and LLM-as-judge approaches on sentence-level hawkish-dovish classification, achieving up to 71.1\% accuracy. Furthermore, the resulting meeting-level stance scores exhibit strong economic relevance. They are closely aligned with real-world macroeconomic conditions, achieving Spearman correlations of up to 0.62 and 0.55 with year-over-year changes in the Consumer Price Index (CPI) and Producer Price Index (PPI), respectively. In addition, the stance scores show highly significant associations with Treasury yields across multiple maturities in regression analyses, indicating that they capture policy signals that are both economically meaningful and reflected in financial market pricing.
Our main contributions are as follows:
\begin{enumerate}
    \item We formalize monetary-policy stance as a relative signal across meetings and propose DCS, the first scoring framework that aligns absolute stance scores with directional shifts between consecutive statements\footnote{Code: https://github.com/yixuantt/DeltaConsistentScoring}.
    \item We demonstrate that DCS, despite requiring no human annotations for training, consistently outperforms both supervised and LLM-as-judge baselines across LLMs of varying scale.
    \item We validate the economic significance of DCS-derived scores by showing strong correlations with inflation indices and significant associations with Treasury yields, confirming that pretrained LLM representations encode actionable monetary-policy information.  Practitioners may adopt our approach to systematically quantify the hawkish–dovish stance embedded in FOMC communication for use in macroeconomic analysis and financial decision-making.
\end{enumerate}

\section{Related Work}

\paragraph{Measuring monetary-policy stance from text.}
Quantitative analysis of central bank communication has evolved through three paradigms. The earliest approaches rely on predefined dictionaries and word-frequency statistics to score FOMC statements~\citep{LuccaDavidO2009MCBC, LOUGHRANTIM2011WIaL}. While transparent and reproducible, these methods count isolated words and miss the discourse-level context. A second paradigm applies machine learning to richer text representations. \citet{HansenStephen2018TADW} use unsupervised topic models on FOMC transcripts. \citet{shah-etal-2023-trillion} construct an expert-annotated dataset of FOMC statements, and show that fine-tuned RoBERTa substantially outperforms dictionary methods. More recently, \citet{Christiano2025From} fine-tune LLMs on a multilingual central bank corpus, and \citet{gambacorta2024cb} benchmark a suite of central-bank language models on FOMC stance labeling. However, these supervised approaches require costly manual annotations that struggle to generalize across evolving rate cycles~\citep{kanganis2025opfed}. The third paradigm leverages LLMs as zero- or few-shot judges. \citet{hansencangpt2023} show that GPT-4 can classify FOMC sentence stance at near-expert level, and \citet{peskoff-etal-2023-gpt} use GPT-4 to quantify within-meeting dissent among hawks and doves. Yet these methods remain sensitive to prompt design and decoding temperature. Across these three paradigms, hawkish--dovish analysis is typically formulated as an isolated classification problem, overlooking the inter-meeting shifts that markets respond to ~\citep{GurkaynakRefetS2005DASL,DohTaeyoung2020DFRC}.

\paragraph{Self-supervised probing of latent representations.}
Our work addresses this gap by building on representation probing, which has shown that pretrained LLMs encode rich semantic concepts in their latent spaces~\citep{alain2018understanding, zou2025representation}. A closely related method is Contrast-Consistent Search (CCS)~\citep{ccs2023}, which discovers latent knowledge by enforcing consistency between a statement and its logical negation. Subsequent work has refined unsupervised modules through spectral methods~\citep{stoehr-etal-2024-unsupervised}, and \citet{park2025steer} showed that steering vectors derived from hidden states can separate truthful from hallucinated outputs. However, these methods are designed for binary distinctions in static settings. Our proposed DCS adapts this line of work in two key ways. First, it replaces the logical negation pair with a chronological pair: two consecutive FOMC statements whose temporal ordering provides a natural contrast. Second, it maps representations to a continuous policy score rather than a binary label, enforcing that the difference between two absolute scores aligns with the relative shift for the same pair. This design turns temporal structure into self-supervision, enabling label-free recovery of continuous monetary-policy stance scores from LLM representations.

\section{Method}
\label{sec:method}
\begin{figure*}[t]
  \centering
  \includegraphics[width=.9\linewidth]{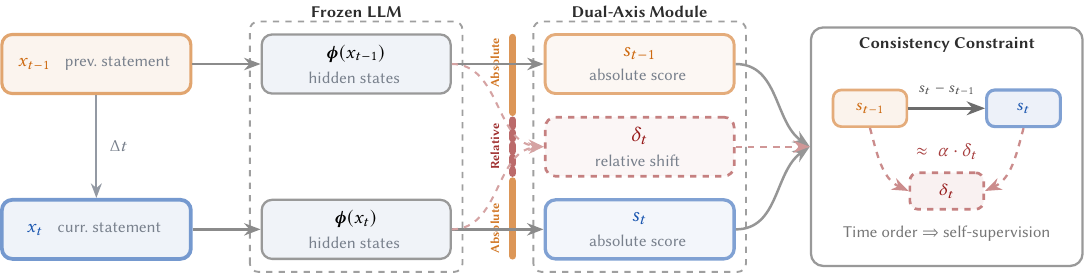}
 \caption{\textbf{Overview of Delta-Consistent Scoring (DCS).} Given two consecutive FOMC statements, a frozen LLM produces absolute and relative representations. A dual-axis projection module maps these representations to an absolute stance score for each statement and estimates the relative shift between them. DCS then aligns score differences with estimated shifts through a delta-consistency objective, turning temporal ordering into a source of self-supervision without requiring human stance labels.}
 \Description{A left-to-right diagram contains four stages: consecutive statements, a frozen language model, a dual-axis scoring module, and a consistency constraint. The module produces absolute scores for the previous and current statements and a relative shift score. Solid arrows trace the absolute-score paths, while dashed arrows trace the relative-shift path. A triangular diagram on the right compares the difference between the absolute scores with the
scaled relative shift.}
  \label{fig:method}
\end{figure*}

We formulate monetary-policy stance detection as learning a continuous stance trajectory over a sequence of statements, rather than as an isolated classification task. We introduce \textbf{Delta-Consistent Scoring (DCS)}, a framework that maps frozen LLM representations to stance scores by constraining them with relative temporal shifts. The method overview is shown in Figure~\ref{fig:method}.

\subsection{Problem Formulation}

Let $D = \{d_1, d_2, \dots, d_T\}$ denote a temporal sequence of FOMC statements. For each statement $d_t$, our goal is to derive a continuous scalar score $s_t \in [0, 1]$. In this continuous spectrum, values approaching $1$ represent a hawkish stance (reflecting tighter monetary policy), while values approaching $0$ represent a dovish stance (reflecting more accommodative policy).

\subsection{Contextual Feature Extraction}

Instead of fine-tuning a model from scratch, we use hidden representations from a frozen LLM, which have been shown to encode rich semantic information. To separate absolute stance from relative movement, we construct two prompt views for each time step. For each statement, we construct two different prompts:

\textbf{Absolute prompt} $p_{\text{abs}}(d_t)$: Asks the LLM to independently assess the absolute policy stance of a single statement $d_t$.
\begin{quote}
\small
\textbf{Instruction:}

Decide whether the stance is tighter (\emph{Hawkish}) or looser (\emph{Dovish}).

Use only stance cues, including inflation versus employment risk, forward guidance, the pace of hikes or cuts, and balance-sheet policy.

Output exactly one token.

\vspace{0.5em}
\textbf{Input statement:}

\texttt{\{text\}}
\end{quote}

\textbf{Relative prompt} $p_{\text{rel}}(d_{t-1}, d_t)$: Explicitly asks the LLM to evaluate the directional policy shift from the preceding meeting $d_{t-1}$ to the current meeting $d_t$.

\begin{quote}
\small
\textbf{Instruction:}

Compare \textbf{only} the stance shift from \emph{Prev} to \emph{Curr}.

Ignore topic changes. Focus only on whether \emph{Curr} implies tighter or looser policy than \emph{Prev}.

Examples of hawkish shift cues include:
higher-for-longer language, faster hikes, and stronger concern about inflation.

Examples of dovish shift cues include:
more accommodation, faster cuts, and stronger concern about employment.

Output exactly one token.

\vspace{0.5em}
\textbf{Previous statement:}

\texttt{\{prev\}}

\vspace{0.5em}
\textbf{Current statement:}

\texttt{\{curr\}}
\end{quote}

We process these prompts through the LLM and extract the hidden states at the last token position of the final layer, yielding text representations $h^{\text{abs}}_t, h^{\text{rel}}_t \in \mathbb{R}^d$.

\subsection{Dual-Axis Projection}

We map the two representations to scalar logits using a lightweight dual-axis projection module:
\begin{align}
    z^{\text{abs}}_t &= \theta_{\text{abs}}^\top h^{\text{abs}}_t + b_{\text{abs}}, \\
    z^{\text{rel}}_t &= \theta_{\text{rel}}^\top h^{\text{rel}}_t + b_{\text{rel}},
\end{align}
where $\theta_{\text{abs}}, \theta_{\text{rel}} \in \mathbb{R}^d$ are learned direction vectors defining the two axes, and $b_{\text{abs}}, b_{\text{rel}} \in \mathbb{R}$ are scalar biases. Here, $z^{\text{abs}}_t \in \mathbb{R}$ is a scalar score representing the absolute policy stance of statement $d_t$, while $z^{\text{rel}}_t \in \mathbb{R}$ is a scalar score representing the policy shift from $d_{t-1}$ to $d_t$. 
The final stance score for a given statement is obtained via a sigmoid mapping: $s_t = \sigma(z^{\text{abs}}_t)$. This continuous score provides a quantitative measure of monetary-policy stance that can be used for downstream economic and financial analyses.

\subsection{Delta-Consistent Objective}

Since DCS is trained without stance annotations, we derive supervision from the temporal structure of FOMC statements. Our core assumption is that the change in absolute stance between two consecutive statements should agree with the relative shift predicted for the same pair. If the current statement becomes more hawkish than the previous one, its absolute stance score should increase; if it becomes more dovish, the score should decrease.

We implement this idea by aligning the change in absolute stance between consecutive statements with the relative shift estimated for the same pair. To prevent extreme relative-shift values from dominating optimization, we bound the relative output with a scaled hyperbolic tangent. The resulting delta-consistency loss is
\begin{equation}
    \mathcal{L}_{\text{delta}} = \mathbb{E}_t \left[
    \left(
    \left(z^{\text{abs}}_t - z^{\text{abs}}_{t-1}\right)
    - \alpha \cdot \tanh\!\left(\frac{z^{\text{rel}}_t}{\tau}\right)
    \right)^2
    \right],
    \label{eq:delta_loss}
\end{equation}
where $\alpha > 0$ is a learnable scale parameter and $\tau$ is a fixed temperature hyperparameter.

The delta-consistency loss provides the main self-supervised signal in our framework, since it directly models the relative changes that are central to monetary-policy communication. However, optimizing this term alone does not produce clear absolute stance scores. We add an auxiliary confidence regularizer that discourages scores near $0.5$ and improves separability along the absolute axis. We minimize the Shannon entropy of the absolute stance predictions:
\begin{equation}
    \mathcal{L}_{\text{conf}} = -\mathbb{E}_t \left[
    s_t \log s_t + (1 - s_t)\log(1 - s_t)
    \right].
\end{equation}

Our final training objective is
\begin{equation}
    \mathcal{L} = \mathcal{L}_{\text{delta}} + \lambda \mathcal{L}_{\text{conf}}.
\end{equation}
Here, $\mathcal{L}_{\text{delta}}$ is the primary training objective, while $\mathcal{L}_{\text{conf}}$ serves as an auxiliary regularizer on the absolute stance scale. During training, we apply a delayed warm-up schedule to $\lambda$. This keeps the confidence regularizer weak in the early stage, allowing the model to first learn the temporal structure of relative stance shifts. As training proceeds, the regularizer is gradually strengthened to sharpen the absolute stance predictions.

\section{Experiments}
\label{sec:experiments}
In this section, we evaluate DCS from both NLP and economic perspectives. We first assess whether its stance scores align with expert sentence-level annotations, and then examine whether the resulting meeting-level scores are economically meaningful through their associations with inflation indicators and Treasury yields.

\subsection{Data}

Our experiments use three types of data: FOMC post-meeting statements, a sentence-level hawkish--dovish benchmark, and external macroeconomic and market indicators for economic validation. DCS is trained only on the unlabeled sequence of meeting-level statements; benchmark labels are used for evaluation.

\paragraph{FOMC statements.}
We apply DCS to a corpus of official FOMC post-meeting statements spanning January 2003 to December 2025, comprising $T = 200$ meetings. For each meeting, we use the full post-meeting statement and apply a rule-based sentence filter to retain policy-relevant content. This step removes boilerplate and procedural text, such as vote tallies, meeting logistics, and recurring administrative language, that is less likely to convey the Committee's monetary-policy stance. 
% Details of the filtering rules are provided in Appendix~\ref{sec:rule_filter}.

\paragraph{Sentence-level benchmark.}
For sentence-level stance evaluation, we use the hawkish--dovish benchmark introduced by \citet{shah-etal-2023-trillion}. We apply DCS directly to individual labeled sentences and evaluate whether the resulting stance scores align with expert annotations.

\paragraph{Macroeconomic and market data.}
For macroeconomic and market validation, we align meeting-level stance scores with inflation and yield data. Monthly CPI and PPI year-over-year changes are obtained from the Federal Reserve Economic Data\footnote{\url{https://fred.stlouisfed.org/}}. For each FOMC meeting, we match the stance score to the next available CPI and PPI release, yielding $N = 199$ matched observations. Treasury yields at the 2-, 10-, and 20-year maturities are obtained from the Federal Reserve H.15 release and matched to FOMC announcement dates on the same day, yielding $N = 191$ observations.

\subsection{Experimental Setup}

% We evaluate DCS across four frozen LLMs ranging from 1B to 14B parameters: Llama-3.2-1B, Qwen3-4B~\citep{qwen3}, Llama-3.1-8B~\citep{llama3}, and DeepSeek-R1-Distill-Qwen-14B~\citep{deepseek}. The full list of hyperparameters is provided in Appendix~\ref{sec:model_backbones}. For each model, we extract final-layer last-token hidden states under both the absolute and relative prompt templates described in Appendix~\ref{sec:appendix_prompts}. The dual-axis projection module is then trained on these frozen representations using the objective in Section~\ref{sec:method}.
We evaluate DCS across four frozen LLMs ranging from 1B to 14B parameters: Llama-3.2-1B, Qwen3-4B~\citep{qwen3}, Llama-3.1-8B~\citep{llama3}, and DeepSeek-R1-Distill-Qwen-14B~\citep{deepseek}. All models use final-layer last-token representations with a maximum input length of 512. We use learning rates of $0.01$ for Llama-3.2-1B and $5\times10^{-4}$ for the other models, train for 1,000--2,000 epochs. The confidence regularizer uses 100 warm-up epochs followed by a 100-epoch linear ramp. The dual-axis projection module is trained on these frozen representations using the objective in Section~\ref{sec:method}.

We evaluate DCS in two settings. For sentence-level evaluation, we apply DCS directly to individual labeled sentences from the benchmark dataset and use the resulting absolute stance scores for classification. For meeting-level analysis, we apply DCS to the full text of each FOMC post-meeting statement and use the resulting absolute stance score as the meeting-level stance measure. Because the learning objective is invariant to a global sign flip, the learned stance axis may be reversed. After training, we use fixed hawkish and dovish exemplars to orient the axis, flipping its sign when necessary. This step does not affect training or the relative score structure; it only makes higher scores interpretable as more hawkish.

\subsection{Evaluation}

We evaluate DCS along two dimensions.

\paragraph{Sentence-level stance classification.}
We evaluate on the hawkish--dovish benchmark introduced by \citet{shah-etal-2023-trillion}, which contains sentence-level annotations from FOMC post-meeting statements. Although DCS is trained without stance labels, this benchmark provides an external evaluation of whether its stance scores align with expert judgment.

\paragraph{Macroeconomic and market validation.}
We further evaluate the economic relevance of the meeting-level stance scores using two validation tests. First, we examine their association with inflation indicators, specifically the CPI and the PPI. Because these indicators are released after the FOMC meeting but reflect prevailing inflation conditions, this analysis assesses whether the stance scores capture the macroeconomic environment underlying policy communication. Second, we examine their association with same-day Treasury yields observed after the statement release. This serves as a market-based validation, as Treasury yields reflect financial market reactions to FOMC communication and changes in expectations about future interest rates.

For inflation, we report Pearson and Spearman correlations between the stance scores and year-over-year CPI and PPI changes. For yields, we match each meeting to the Treasury yield observed after the FOMC statement release on the same announcement date, and estimate
\begin{equation}
    y_t = \alpha + \beta \, \tilde{s}_t + \varepsilon_t,
\end{equation}
where $y_t$ is the Treasury yield on the FOMC announcement date and $\tilde{s}_t$ is the standardized stance score with zero mean and unit variance. Standard errors are computed using the Newey--West estimator to account for serial correlation.

\subsection{Baselines}

We compare DCS against several baselines spanning dictionary-based, supervised, and prompt-based approaches to monetary-policy stance detection.

The first baseline is a dictionary method, which assigns each input a lexicon-based score computed as the difference between hawkish and dovish word counts~\citep{GorodnichenkoYuriy2023TVoM}. This baseline represents the traditional keyword-matching approach to monetary-policy stance measurement.

The second baseline is a supervised benchmark model from prior work. Specifically, we report the performance of the RoBERTa-based supervised model released by \citet{shah-etal-2023-trillion}, which is trained on expert-annotated hawkish and dovish sentence labels. In the sentence-level evaluation, we apply the model directly to individual sentences. In the meeting-level evaluation, we follow \citet{shah-etal-2023-trillion} and aggregate sentence-level predictions into a normalized meeting-level stance score. 

For each frozen LLM backbone, we further compare DCS against two model-specific baselines. The first is an LLM Judge baseline~\citep{hansencangpt2023}, which prompts the model to directly output a hawkish or dovish judgment in natural language. The second is a logit-based LLM Judge baseline, which derives a stance signal from the model's next-token preferences for the labels Hawkish and Dovish under the absolute prompt. We also include a supervised linear probe baseline, which trains a linear classification head on top of the same frozen LLM representations using the training split of the expert-annotated data from \citet{shah-etal-2023-trillion}. This comparison isolates whether DCS improves over standard supervised probing on the same backbone.

\subsection{Main Results}

We first evaluate DCS on sentence-level hawkish--dovish classification, and then examine whether the resulting meeting-level stance scores are economically meaningful.

\begin{table}[t]
\centering
\caption{Sentence-level stance classification results. Best result in each metric is \textbf{bolded}.}
\label{tab:sentence_results}
\footnotesize
\setlength{\tabcolsep}{5pt}
\resizebox{.9\linewidth}{!}{
\begin{tabular}{llcc}
\toprule
Model & Method & Acc. & F1 \\
\midrule
-- & Dictionary & 0.4458 & 0.5893 \\
FOMC-RoBERTa & Supervised & 0.4337 & 0.5346 \\
\midrule
\multirow{4}{*}{Llama-3.2-1B}
& LLM Judge & 0.5060 & 0.5393 \\
& Logit-Based Judge & 0.5301 & 0.6929 \\
& Linear Probe & 0.5542 & 0.6667 \\
& DCS (Ours) & 0.5904 & 0.7119 \\
\midrule
\multirow{4}{*}{Qwen3-4B}
& LLM Judge & 0.6386 & 0.6939 \\
& Logit-Based Judge & 0.5783 & 0.6316 \\
& Linear Probe & 0.6145 & 0.6190 \\
& DCS (Ours) & \textbf{0.7108} & 0.7333 \\
\midrule
\multirow{4}{*}{Llama-3.1-8B}
& LLM Judge & 0.5422 & 0.6415 \\
& Logit-Based Judge & 0.5301 & 0.6929 \\
& Linear Probe & 0.5904 & 0.6222 \\
& DCS (Ours) & 0.6024 & 0.6374 \\
\midrule
\multirow{4}{*}{DeepSeek-R1-14B}
& LLM Judge & 0.5301 & 0.6929 \\
& Logit-Based Judge & 0.5301 & 0.6929 \\
& Linear Probe & 0.5542 & 0.6186 \\
& DCS (Ours) & 0.6506 & \textbf{0.7387} \\
\bottomrule
\end{tabular}
}
\end{table}

\paragraph{Sentence-level stance classification.}
Table~\ref{tab:sentence_results} reports sentence-level hawkish--dovish classification results. Across all four LLM backbones, DCS outperforms all corresponding baselines in accuracy, including the LLM Judge, Logit-Based Judge, and the supervised Linear Probe. The best accuracy of 71.1\% is achieved by DCS on Qwen3-4B, and the best F1 of 0.74 by DCS on DeepSeek-R1-14B. Both the Dictionary baseline and the supervised FOMC-RoBERTa model fall below 45\% accuracy, suggesting that keyword counting and prior supervised approaches struggle on this task. Among the model-specific baselines, the LLM Judge and Logit-Based Judge perform comparably to or below the Linear Probe in most cases, indicating that prompting alone does not reliably elicit stance judgments from frozen LLMs. DCS improves over the Linear Probe by margins ranging from about 1 percentage point (Llama-3.1-8B) to nearly 10 percentage points (Qwen3-4B and DeepSeek-R1-14B) in accuracy, despite requiring no stance labels during training.

\begin{table}[htbp]
\centering
\caption{Pearson and Spearman correlations between meeting-level stance scores and year-over-year inflation changes. Best result in each column is \textbf{bolded}.}
\label{tab:full_correlation}
\resizebox{\linewidth}{!}{
\begin{tabular}{llcccc}
\toprule
 & & \multicolumn{2}{c}{CPI (YoY)} & \multicolumn{2}{c}{PPI (YoY)} \\
\cmidrule(lr){3-4} \cmidrule(lr){5-6}
Model & Method & Pearson & Spearman & Pearson & Spearman \\
\midrule
-- & Dictionary & 0.3137 & 0.3837 & 0.2894 & 0.2988 \\
FOMC-RoBERTa & Supervised & 0.3880 & 0.4458 & 0.2884 & 0.2849 \\
\midrule
Llama-3.2-1B & LLM Judge & 0.2375 & 0.3190 & 0.1516 & 0.1799 \\
Llama-3.2-1B & Logit-Based Judge & 0.0653 & 0.2271 & 0.0597 & 0.1810 \\
Llama-3.2-1B & Linear Probe & 0.1960 & 0.4039 & 0.1554 & 0.3755 \\
Llama-3.2-1B & DCS (Ours) & 0.3447 & 0.5402 & 0.3884 & 0.5085 \\
\midrule
Qwen3-4B & LLM Judge & 0.1699 & 0.1414 & 0.0800 & 0.0484 \\
Qwen3-4B & Logit-Based Judge & 0.3796 & 0.3411 & 0.2531 & 0.2108 \\
Qwen3-4B & Linear Probe & 0.1659 & 0.2058 & 0.0317 & 0.0091 \\
Qwen3-4B & DCS (Ours) & 0.4590 & 0.6055 & 0.4337 & 0.5236 \\
\midrule
Llama-3.1-8B & LLM Judge & 0.3538 & 0.2803 & 0.2170 & 0.1964 \\
Llama-3.1-8B & Logit-Based Judge & \textbf{0.5034} & 0.4166 & 0.3295 & 0.2924 \\
Llama-3.1-8B & Linear Probe & 0.0341 & 0.1616 & 0.0000 & 0.1869 \\
Llama-3.1-8B & DCS (Ours) & 0.4809 & 0.5382 & 0.4244 & 0.4546 \\
\midrule
DeepSeek-R1-14B & LLM Judge & 0.3888 & 0.3397 & 0.2573 & 0.2424 \\
DeepSeek-R1-14B & Logit-Based Judge & 0.2268 & 0.1793 & 0.0778 & 0.0685 \\
DeepSeek-R1-14B & Linear Probe & 0.0547 & 0.0720 & 0.0607 & 0.0389 \\
DeepSeek-R1-14B & DCS (Ours) & 0.5021 & \textbf{0.6237} & \textbf{0.4799} & \textbf{0.5530} \\
\bottomrule
\end{tabular}
}
\end{table}

\begin{table}[htbp]
\centering
\caption{Regression of Treasury yield levels on standardized stance scores. $\beta$ denotes the slope coefficient with Newey--West standard errors; $p$ is the corresponding $p$-value. }
\label{tab:yield_level_regression_scores}
\resizebox{\linewidth}{!}{
\begin{tabular}{llcccccc}
\toprule
 & & \multicolumn{2}{c}{2Y Yield} & \multicolumn{2}{c}{10Y Yield} & \multicolumn{2}{c}{20Y Yield} \\
\cmidrule(lr){3-4} \cmidrule(lr){5-6} \cmidrule(lr){7-8}
Model & Method & $\beta$ & $p$ & $\beta$ & $p$ & $\beta$ & $p$ \\
\midrule
FOMC-RoBERTa & Supervised & 1.058 & {$<$.01} & 0.536 & {$<$.01} & 0.402 & {$<$.01} \\
\midrule
Llama-3.1-8B & LLM Judge & 0.854 & {$<$.01} & 0.328 & {$<$.01} & 0.202 & .096 \\
Llama-3.1-8B & Logit-Based Judge & 0.901 & {$<$.01} & 0.299 & .012 & 0.158 & .185 \\
Llama-3.1-8B & Linear Probe & 0.055 & .754 & 0.227 & .067 & 0.326 & {$<$.01} \\
Llama-3.1-8B & DCS (Ours) & 0.842 & {$<$.01} & 0.796 & {$<$.01} & 0.852 & {$<$.01} \\
\midrule
DeepSeek-R1-14B & DCS (Ours) & 2.121 & {$<$.01} & 0.782 & {$<$.01} & 0.771 & {$<$.01} \\
\bottomrule
\end{tabular}
}
\end{table}

\paragraph{Macroeconomic validation: CPI and PPI.} 
To evaluate the economic relevance of the stance measure, we examine whether meeting-level stance scores are associated with inflation conditions captured by year-over-year changes in CPI and PPI. Table~\ref{tab:full_correlation} reports Pearson and Spearman correlations between the stance scores and year-over-year changes in CPI and PPI. Overall, DCS produces the strongest and most consistent results across backbones, with especially large gains under Spearman correlation. The best overall results are obtained with DeepSeek-R1-14B, which reaches a Spearman correlation of 0.6237 for CPI and 0.5530 for PPI. DCS also achieves the strongest Pearson correlation for PPI (0.4799), while remaining competitive on CPI Pearson correlation across models.

This pattern is broadly consistent across backbones. Under Spearman correlation, DCS attains CPI correlations between 0.5382 and 0.6237 and PPI correlations between 0.4546 and 0.5530, consistently outperforming the LLM Judge, Logit-Based Judge, and Linear Probe built on the same backbone. In contrast, the supervised FOMC-RoBERTa baseline reaches 0.4458 on CPI and 0.2849 on PPI, remaining below the strongest DCS variants. Linear Probe results are generally weaker and less stable, despite using supervised sentence-level labels. Additional OLS regression results in Table~\ref{tab:appendix_regression} further support the positive association between DCS stance scores and inflation outcomes.
Because the matched CPI and PPI releases occur after the FOMC meeting, these correlations suggest that the estimated stance scores capture prevailing inflation conditions and provide a quantitative measure of the macroeconomic environment.

\paragraph{Regression Analysis of Inflation}

As an additional economic validation, we fit separate OLS regressions of year-over-year CPI and PPI changes on meeting-level stance scores produced by DCS with the DeepSeek-R1-Distill-Qwen-14B backbone. The estimated association is positive and statistically significant for both indicators. For CPI, a one-unit increase in the stance score is associated with a 3.18 percentage-point increase in year-over-year CPI growth ($p<0.0001$), with $R^2=0.2521$ over $N=199$ observations. For PPI, the corresponding coefficient is 12.77 percentage points ($p<0.0001$), with $R^2=0.2303$. These results are consistent with the correlation analysis and suggest that higher DCS scores are associated with stronger inflation conditions. We interpret these estimates as associations rather than causal effects.

\begin{table}[htbp]
\centering
\small
\caption{OLS regressions of year-over-year CPI and PPI changes on meeting-level DCS stance scores. Standard errors are shown in parentheses. *** $p<0.01$.}
\label{tab:appendix_regression}
\setlength{\tabcolsep}{6pt}
\begin{tabular}{lcc}
\toprule
 & CPI change & PPI change \\
\midrule
Intercept & 0.75*** & -3.05*** \\
 & (0.23) & (1.15) \\
DCS score & 3.18*** & 12.77*** \\
 & (0.56) & (2.45) \\
\midrule
$N$ & 199 & 199 \\
$R^2$ & 0.2521 & 0.2303 \\
\bottomrule
\end{tabular}
\end{table}

\paragraph{Market validation: Treasury yields.} Same-day Treasury yields provide a test of whether the extracted stance scores capture the hawkish–-dovish policy signal in FOMC statements that financial markets incorporate into interest rate expectations.  Table~\ref{tab:yield_level_regression_scores} reports regressions of Treasury yield levels on standardized stance scores at the 2-, 10-, and 20-year maturities. Among the Llama-3.1-8B baselines, the LLM Judge and Logit-Based Judge are significant only at shorter maturities, while the Linear Probe is mostly insignificant. In contrast, DCS on the same backbone remains significant across all three maturities. The strongest results come from DeepSeek-R1-14B, where DCS reaches $\beta = 2.12$ at 2 years and remains significant at 10 and 20 years. Overall, DCS is the only method that shows consistently significant associations with Treasury yields across maturities. These results suggest that DCS captures the market-relevant policy signal in FOMC statements more effectively than the baseline methods.

\section{FOMC Stance Trajectory and Robustness Across Policy Regimes}

\begin{figure*}[htbp]
\centering
\includegraphics[width=.8\linewidth]{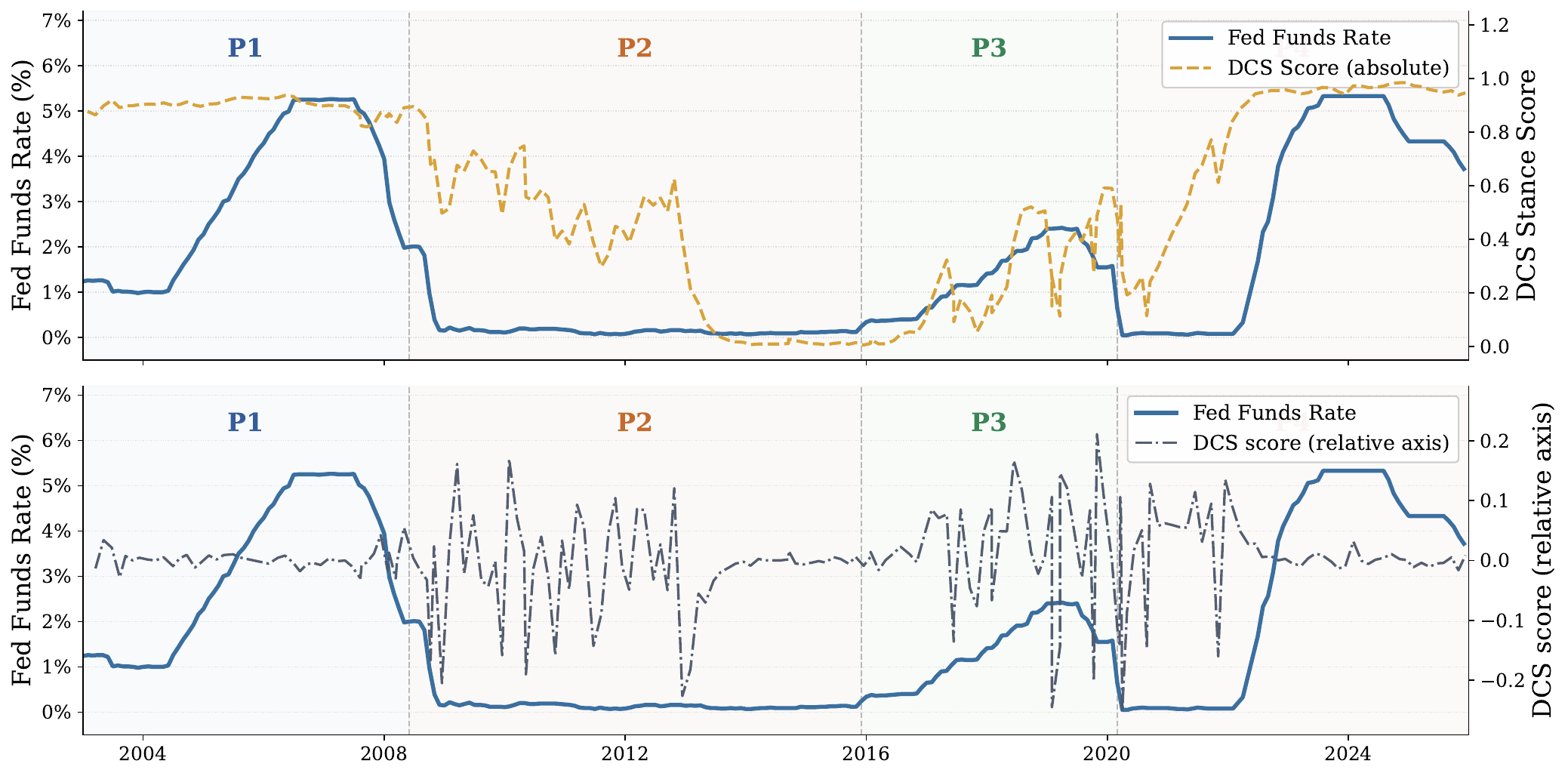}
\caption{Federal Funds Rate (blue, left axis) and DCS scores over time. The top panel shows stance scores from the absolute axis (orange dashed, right axis), which are used in the main analyses, while the bottom panel shows scores from the relative axis (gray dash-dotted, right axis). Vertical dashed lines divide the sample into four policy periods, with shaded regions marking P1 (2003--2008, conventional policy), P2 (2008--2015, near-zero rates), P3 (2015--2020, policy normalization), and P4 (2020--2025, pandemic and tightening cycle). }
\Description{Two vertically stacked time-series panels cover 2003 to 2025. Both panels plot the Federal Funds Rate as a blue solid line against the left axis. The upper panel adds absolute DCS stance scores as an orange dashed line against the right axis. The lower panel adds relative DCS shift scores as a gray dash-dotted line against the right axis.
Vertical dashed boundaries and shaded regions mark four periods: conventional policy, near-zero rates, policy normalization, and the pandemic and tightening cycle.}
\label{fig:policy_periods}
\end{figure*}

\begin{table}[t]
\centering
\caption{Temporal robustness across policy regimes on the DeepSeek-R1-Distill-Qwen-14B backbone. We report period-specific Pearson ($r$) and Spearman ($\rho$) correlations between meeting-level stance scores and subsequent inflation indicators.}
\label{tab:period_robustness_deepseek}
\resizebox{\linewidth}{!}{%
\begin{tabular}{llcccc}
\toprule
 & & P1 & P2 & P3 & P4 \\
\cmidrule(lr){3-6}
Inflation & Method & 2003--2008 & 2008--2015 & 2015--2020 & 2020--2025 \\
\midrule
\multirow{4}{*}{CPI ($r/\rho$)}
& LLM Judge & 0.282 / 0.275 & 0.052 / -0.017 & 0.276 / 0.260 & 0.377 / 0.440 \\
& Logits Based Judge & 0.435 / 0.537 & -0.118 / -0.107 & 0.104 / 0.195 & 0.286 / 0.325 \\
& Linear Probe & 0.027 / 0.035 & -0.040 / -0.087 & 0.009 / 0.152 & 0.225 / 0.039 \\
& DCS (Ours) & 0.269 / 0.082 & 0.108 / 0.076 & 0.339 / 0.354 & 0.376 / 0.199 \\
\midrule
\multirow{4}{*}{PPI ($r/\rho$)}
& LLM Judge & 0.154 / 0.067 & 0.013 / -0.029 & 0.350 / 0.301 & 0.441 / 0.376 \\
& Logits Based Judge & 0.310 / 0.487 & -0.149 / -0.168 & 0.003 / 0.074 & 0.232 / 0.266 \\
& Linear Probe & 0.029 / -0.008 & 0.055 / -0.107 & 0.039 / 0.137 & 0.377 / 0.024 \\
& DCS (Ours) & 0.273 / 0.056 & 0.228 / 0.228 & 0.158 / 0.184 & 0.686 / 0.617 \\
\bottomrule
\end{tabular}%
}
\end{table}

We estimate the FOMC stance trajectory over time and assess its robustness by partitioning the sample into four periods defined by major shifts in Federal Reserve policy. Figure~\ref{fig:policy_periods} illustrates the four policy periods used in our robustness analysis. Period 1 (2003--2008) covers the pre-crisis era of conventional monetary policy. Period 2 (2008--2015) begins with the Fed's move to near-zero interest rates following the 2008 financial crisis, a regime in which standard rate signals largely disappeared. Period 3 (2015--2020) covers the gradual return to normal rate levels. Period 4 (2020--2025) spans the pandemic-induced rate cuts and the sharp inflation-driven tightening cycle that followed. These breakpoints mark well-recognized discontinuities in how the Fed communicates policy, and thus provide a natural stress test for stance-scoring methods.\footnote{Breakpoints follow official Federal Reserve descriptions of major policy transitions.}

Table~\ref{tab:period_robustness_deepseek} reports the results. DCS is the only method that maintains positive correlations with both CPI and PPI across all four periods under both Pearson and Spearman metrics. The supervised Linear Probe is the weakest baseline: its correlations remain near zero in Periods 1--3 and become unstable in Period 4.

Among the prompt-based baselines, the Logits Based Judge performs strongly in the pre-crisis period, achieving the highest Period~1 correlations for both CPI ($r=0.435$, $\rho=0.537$) and PPI ($r=0.310$, $\rho=0.487$), but turns negative for both indicators in Period~2, suggesting its score boundaries are sensitive to the low-rate regime. The LLM Judge is more stable but still produces near-zero or negative Spearman correlations in Period~2. DCS, by contrast, remains positive throughout and achieves the strongest correlations in the 2020--2025 tightening period for PPI ($r=0.686$, $\rho=0.617$).

We also plot the relative-axis score trajectory in the bottom panel of Figure~\ref{fig:policy_periods}, which are designed to capture meeting-to-meeting shifts in policy stance. Notably, the relative DCS scores fluctuate more sharply during the post–financial crisis period characterized by quantitative easing (QE), as well as around the onset of the COVID-19 pandemic. Both episodes correspond to substantial shifts in the Federal Reserve’s policy framework and communication strategy. This pattern suggests that the relative component effectively captures policy transitions during periods of unconventional monetary policy and macroeconomic stress.

\section{Ablation Study}

\begin{table}[t]
\centering
\caption{Ablation study on DeepSeek-R1-14B. Acc.\ and F1 are evaluated on the hawkish--dovish benchmark. CPI and PPI columns report Spearman correlations with year-over-year inflation changes.}
\label{tab:ablation}
\small
\begin{tabular}{lcccc}
\toprule
Variant & Acc. & F1 & CPI ($\rho$) & PPI ($\rho$) \\
\midrule
Full DCS & 0.6506 & 0.7387 & 0.6237 & 0.5530 \\
\quad w/o $\mathcal{L}_{\text{delta}}$ & 0.4699 & 0.5000 & 0.3807 & 0.3133 \\
\quad w/o $\mathcal{L}_{\text{conf}}$ & 0.5904 & 0.7213 & 0.2545 & 0.1521 \\
\quad Single axis & 0.5867 & 0.6436 & 0.5498 & 0.4169 \\
\bottomrule
\end{tabular}
\vspace{-1.5em}
\end{table}

\paragraph{Loss components and architecture.}
We test three core components of DCS on DeepSeek-R1-14B: the delta-consistency loss $\mathcal{L}_{\text{delta}}$, the confidence regularizer $\mathcal{L}_{\text{conf}}$, and the dual-axis design. Table~\ref{tab:ablation} reports the results.  Removing $\mathcal{L}_{\text{delta}}$ causes the largest drop: accuracy falls to 0.47, and the CPI correlation drops from 0.62 to 0.38, confirming that it is the primary learning signal. Removing $\mathcal{L}_{\text{conf}}$ preserves sentence-level F1 (0.72) but reduces the CPI correlation to 0.25, indicating that the regularizer is essential for calibrating the absolute stance scale. In the single-axis variant, both projections use the absolute-prompt representation, while the two projections are still trained separately. This variant lowers all four metrics, showing that separating absolute and relative representations helps the model better capture each signal. 

\begin{figure}[t]
  \centering
  \includegraphics[width=\linewidth]{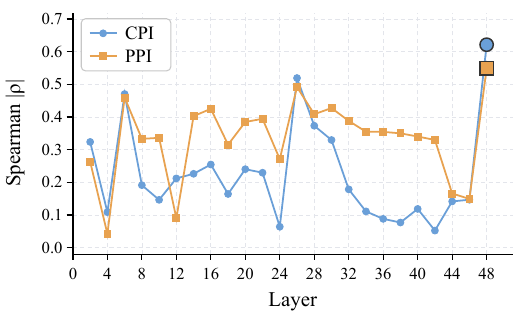}
  \caption{Spearman $\rho$ between DCS stance scores and year-over-year CPI and PPI changes across layers of DeepSeek-R1-14B. The final layer (L48, highlighted) achieves the highest correlations on both indicators.}
  \Description{A line chart compares inflation correlations across sampled layers from L2 to L48. Blue circles represent CPI and orange squares represent PPI. Both series vary across intermediate layers, with a shared local peak at L26. The final layer, L48, is highlighted with larger outlined markers and has the highest correlations: 0.62 for CPI and 0.55 for PPI.}
  \label{fig:layer_sweep}
\end{figure}

\paragraph{Layer selection.}
\label{sec:layer_analysis}
We compare DCS performance across different layers of DeepSeek-R1-14B to validate the choice of using the final layer. We extract hidden states from every other layer (L2, L4, \dots, L46) as well as the final layer (L48), and train a separate DCS module on each. Figure~\ref{fig:layer_sweep} shows the results. The final layer achieves the strongest correlations on both indicators ($\rho = 0.62$ for CPI, $0.55$ for PPI), outperforming all intermediate layers by a clear margin. The second-best results come from L26 ($\rho = 0.52$ for CPI, $0.49$ for PPI). Performance varies substantially across layers, with many early and late-middle layers yielding weak correlations.

\section{Conclusion}
In this work, we study how LLMs can be used to decode monetary-policy signals from FOMC communication.  We introduced Delta-Consistent Scoring (DCS), an annotation-free framework that maps frozen LLM representations to continuous stance scores by aligning absolute scores with relative shifts between consecutive FOMC meetings. Across four LLMs, DCS consistently outperforms supervised and prompt-based baselines on sentence-level classification, and produces meeting-level scores that correlate strongly with inflation indicators and Treasury yields. Ablation experiments confirm that each component contributes, with the delta-consistency loss serving as the primary learning signal. These results suggest that frozen LLM representations encode rich monetary-policy information and that temporal self-supervision provides an effective way to extract it. Broadly, this work provides practitioners with a systematic way to extract an economically meaningful and continuous measure of monetary policy stance from FOMC communication for macroeconomic analysis and financial decision-making.

\bibliographystyle{ACM-Reference-Format}
\bibliography{custom}

\end{document}